# Using Intelligent Agents to understand organisational behaviour

**Objectives**

In this paper we describe two research projects, whose objectives are to: (1) bring together expertise in organisational psychology, computer science, and organisational practice, (2) address a set of applied problems, (3) simulate human behaviour in different organisational situations, and (4) develop and exploit long-term capability in this area.

**Background**

Researchers across a range of disciplines have for some time attempted to understand the behaviour of complex systems using simulation techniques. Operations researchers, for example, have long-standing interests in flows of traffic in cities and the movement of materials through factories and supply chains. Such simulation techniques employ Intelligent Agents – computer systems capable of autonomous problem solving – using an approach known as Agent Based Modelling and Simulation (ABMS). ABMS is used to study how micro-level processes affect macro-level outcomes; macro behaviour is not simulated, rather it emerges from the micro decisions made by individual agents.

Research within biology has demonstrated that apparently complex behaviour on the part of insects and animals can be simulated using intelligent agents which are programmed to follow simple rules. (See Gilbert & Troitzsch, 2005 for a summary of the approach). For example, scientists have simulated two different kinds of food-seeking behaviour by ants (Bonabeau, 2002). In the first, labelled 'mass recruitment', each ant leaves the nest randomly seeking food, laying down a pheromone trail as it does so. The first ant to find food returns to the nest, thereby reinforcing the trail. Other ants pick up the trail and copy the route, thereby further reinforcing the trail. Ants keep using the trail until the food runs out. In the second model, labelled 'tandem recruitment', each ant leaves the nest seeking food. On finding food, an ant returns to the nest and recruits one other ant by vibrating its antennae. Both ants return to the food, and then back to the nest where further pair recruitment takes place. They repeat this process until the food runs out. It appears that both methods of food-seeking behaviour have evolved and occur in the wild. The first method is ideal for large sources of food (such as animal corpses), whilst the second is better suited to smaller sources.

The intellectual question driving our research is "can intelligent agents be used to simulate and understand human behaviour in organisations?"

# The Projects

Project 1 is focused on productivity in retail companies in the UK and is funded by the EPSRC. Key partners in the project are the Universities of Leeds and Nottingham. More specifically we are trying to simulate the behaviour of staff and shoppers in a major department store. In this model there are 3 different kinds of agent representing customers, sales staff, and department managers, each behaving and interacting according to certain rules. For example, customers who are browsing behave differently from those who have specific shopping needs in mind. Sales staff differ according to age, skill level, and their attitudes to customers.  (See Siebers, 2006).

Project 2 is concerned with relationships within and between engineering teams and is funded by the DTI and Rolls-Royce. Key partners in the project are Rolls-Royce, Jaguar, and the Universities of Leeds and Southampton. Here we are trying to simulate the ways in which different teams of engineers working on the same project interact with one another, using case studies in each company. For example, when 2 teams (A and B) need to communicate with one another, should everyone in team A be able to talk to everyone in team B (any to any), or should communications from team A be channelled through a single individual and thence to a single individual in team B, who then communicates to all in team B (any to 1 to 1 to any)? We will be modeling different patterns of communication trying to establish their effectiveness under different situations. Variables include shared understanding, trust, leadership, team roles, and effectiveness.

**Approach and Methods**

Both projects involve the following stages: (1) understand the particular problem domain; (2) generate the underlying rules currently in place; (3) simulate the existing system using intelligent agents; (4) test that the simulation is a sufficiently good representation; (5) generate new scenarios for how the system could work using new rules; (6) simulate the new ways of working using intelligent agents; (7) use the outputs of the simulation to recommend new ways of working to improve aspects of system performance. In project 1 we are currently working in stages (1) and (2), whilst in project 2 we are working in stage (1). In both studies we are using a number of methods including observation, questionnaires, interviews, and analysis of company data.

**Conclusions**

We regard both projects as relatively high risk in that they may not add value in the short term. However, we do believe that organisational psychologists and others should become involved in this approach to understanding behaviour in organisations. In our view the main potential benefits from adopting this approach are that it: (1) improves understanding of, and debate about, a problem domain, (2) forces researchers to be explicit about the rules underlying behaviour, (3) opens up new choices and opportunities for ways of working, (4) allows testing of alternatives cheaply, (5) improves the quality of information for those designing / managing systems, and (6) reduces risks. Time will tell if such benefits can be delivered.